\documentclass[conference]{IEEEtran}
\IEEEoverridecommandlockouts
\usepackage{cite}
\usepackage{amsmath,amssymb,amsfonts}
\usepackage{algorithmic}
\usepackage{graphicx}
\usepackage{textcomp}
\usepackage{xcolor}
\usepackage{subcaption}
\usepackage{caption}
\usepackage{amssymb}
\def\BibTeX{{\rm B\kern-.05em{\sc i\kern-.025em b}\kern-.08em
    T\kern-.1667em\lower.7ex\hbox{E}\kern-.125emX}}

\makeatletter
\newcommand{\linebreakand}{%
  \end{@IEEEauthorhalign}
  \hfill\mbox{}\par
  \mbox{}\hfill\begin{@IEEEauthorhalign}
}
\makeatother
\begin{document}

\title{Design Activity for  Robot Faces: Evaluating Child Responses To Expressive Faces\\
}

\author{ Denielle Oliva$^1$, Joshua Knight$^2$, Tyler Becker$^3$, Heather Aministani $^4$, Monica Nicolescu$^5$, David Feil-Seifer$^6$ 
\thanks{$^1$Denielle Oliva, $^2$ Joshua Knight, $^3$ Tyler Becker, $^4$ Heather Amistani, $^5$ Monica Nicolescu and $^6$ David Feil-Seifer are with the Department of Computer Science and Engineering, University of Nevada, Reno, 1664 N. Virginia Street, Reno, NV 89557-0171, USA \emph{denielleo@unr.edu}, \emph{joshuaknight@unr.edu}, \emph{tbecker@unr.edu}, \emph{hamistani606@unr.edu}, \emph{monica@unr.edu} and \emph{dfseifer@unr.edu}}}

\maketitle

\begin{abstract}
Facial expressiveness plays a crucial role in a robot’s ability to engage and interact with children. Prior research has shown that expressive robots can enhance child engagement during human-robot interactions. However, many robots used in therapy settings feature non-personalized, static faces designed with traditional facial feature considerations, which can limit the depth of interactions and emotional connections. Digital faces offer opportunities for personalization, yet the current landscape of robot face design lacks a dynamic, user-centered approach. Specifically, there is a significant research gap in designing robot faces based on child preferences. Instead, most robots in child-focused therapy spaces are developed from an adult-centric perspective. 
We present a novel study investigating the influence of child-drawn digital faces in child-robot interactions. This approach focuses on a design activity with children instructed to draw their own custom robot faces. We compare the perceptions of social intelligence (PSI) of two implementations: a generic digital face and a robot face, personalized using the user's drawn robot faces. The results of this study show the perceived social intelligence of a child-drawn robot was significantly higher compared to a generic face. 

\end{abstract}


\section{Introduction}

% robots in a SLP/learning context

    % robots are being using in therapeutic interactions with children (characterize those environments as closely as possible to what we've been discussing the SLP context)

Face design is a crucial consideration for socially assistive robots~\cite{feil-seifer2005defining} in a therapeutic educational setting. Expressiveness in clinical settings can help communicate social nuances, and can also serve as a supplement for various intervention targets. For example, in speech language pathology spaces, a therapist will often use implicit communication (expressions, hand gestures, etc.) that increases engagement and understanding when a patient is learning different target behaviors or lacks the social understanding to read nuanced information.

    % role of the robot in these settings
A robot in clinical and educational spaces can aid in learning, providing social companionship, or supporting a person providing a clinical or therapeutic service. 
    % effective and accurate expressive communication is crucial to the robot's success in this domain
In order to sustain these interactions in as productive a fashion as possible, a robot has to effectively employ verbal and nonverbal behaviors, and facial expressions are a critical component of such behaviors. For human-to-human relationships, effective communication utilizing implicit and explicit strategies can be key in developing trust and social rapport with people. People can do this by accompanying our dialogue with tone changes, hand gestures, facial expressions, etc.

\begin{figure}[t]
    \centering
    \includegraphics[width=0.85\linewidth]{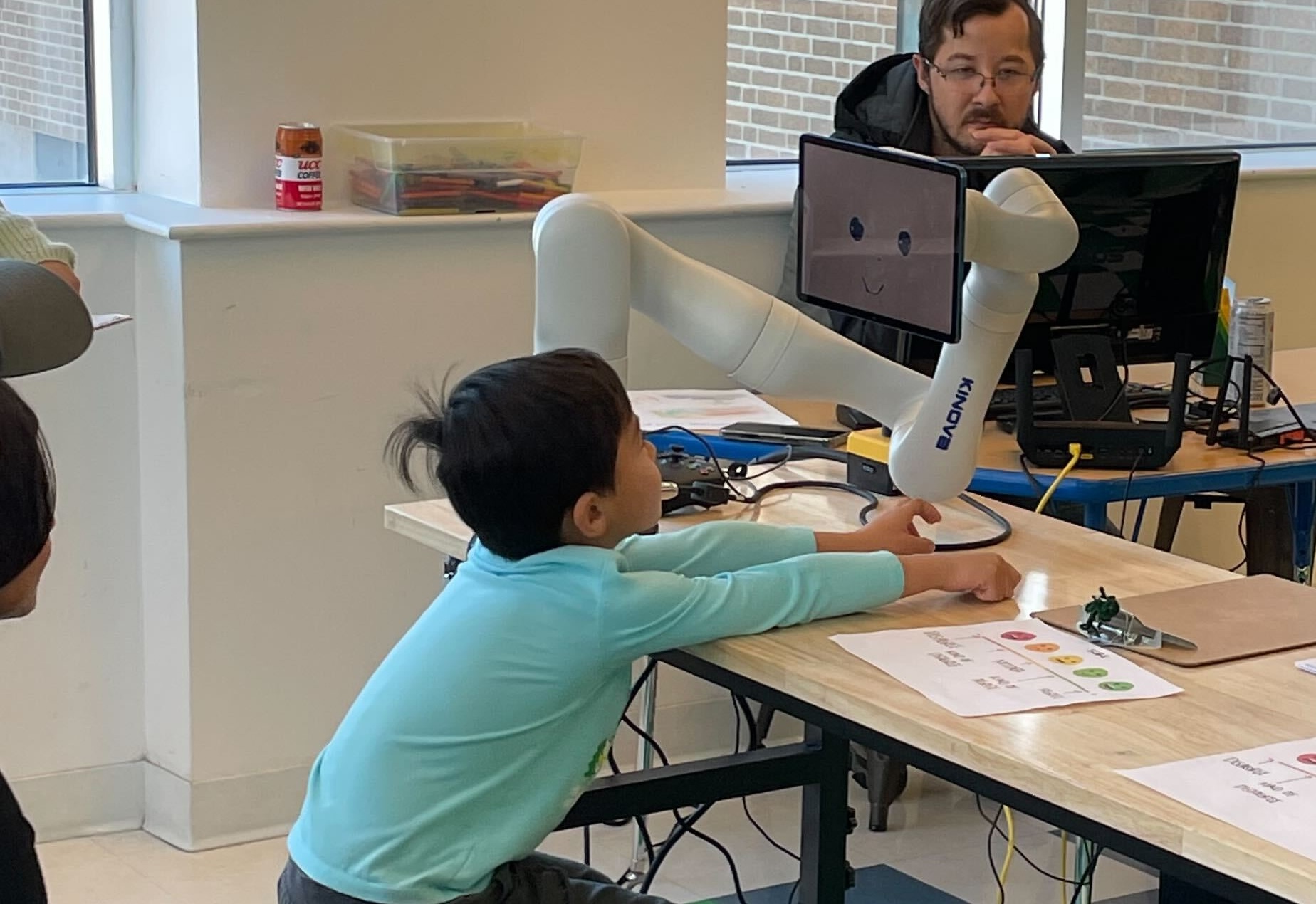}
    \caption{The study setup includes a Kinova Gen3 robot arm with a mounted digital screen. Participants were asked to observe the robot embodiment and answer questions after an introductory behavior.}
    \label{fig:robotsetup}
\end{figure}

% robot ability to communicate verbally and nonverbally is part of effective interaction
 
    % importantance of verbal interaction
    
Effective communication in clinical spaces often involves effective verbal communication. The primary mode of communication in child-robot settings is verbal interaction~\cite{grigore2016talk}. Such communication can help explicitly direct the interaction toward specific goals, but it can also help shape a user's impression of a robot as well. Verbal communication can have an effect on the accumulation or dissolution of trust~\cite{alhaji2021trust}. Verbal communication can also have an effect on perceived social distance with a robot~\cite{kim2013acceptable}. Verbal interaction is further supported by nonverbal interaction.

    % components of effective nonverbal interaction

For robots, conveying emotion through non-verbal cues such as facial expressions is crucial to being an effective agent in child-robot interaction. Brezeal, et al.~\cite{breazeal2003emotion} and Bethel, et al.~\cite{bethel2010affective} highlight this, stating that facial expressions are important for building trust and emotional connections between assistive robots and children. Importantly, these nonverbal gestures can complement verbal interactive behavior, adding richness or emphasis to those cues~\cite{busso2004analysis}.
% role of faces in expressive interaction
Solutions for expressive robots are traditionally implemented using systems with physical faces or digital faces which can limit the flexibility of a system to employ expressiveness as a tool for enhancing communication.

In this paper, we explore a novel approach to designing robot faces for the purpose of interacting with children through personalized faces. Personalization in terms of unique faces offers a promising solution to building deeper connections with children with socially assistive robots. Personalization in HRI has traditionally focused on personalizing behaviors or care regimes; we take the perspective of using a user-created agent and implementing it in an assistive robot.

This paper details a design activity in which child-drawn robot faces are implemented on a digital face robot. The study is further used to determine how the perceived social intelligence of a non-personalized face compares to a personalized face that participants drew, which allows us to provide a quantitative measure of how the different agents performed. Participants were asked if the perceived social intelligence of a child-drawn and generic agent differs in a simple interaction. The results of the study show a significant improvement from a generic face to a child-drawn face.

The remainder of this paper is organized as follows. Section \ref{RelatedWorks} describes the related work, Section \ref{Methods} presents the study design and software used, Section \ref{Results} reports on the results of the study, Section \ref{Discussion} discusses our thoughts on the results of the work presented and \ref{Limitations} describes the limitations that this study did not address.

\section{Related Work} \label{RelatedWorks}

\subsection{Robots in Education and Therapy}

% robots in education and classroom setting

Robot assistants represent a way to reduce the workload of therapists and teachers by providing personal tutoring to children, which would go beyond the time a therapist might be able to give. The work of Serholt et al. \cite{serholt2014} and Westlund et al. \cite{korywestlund2016} shows that teachers are interested in having robots enter the classroom, but are wary that the robot might not only distract children but even increase the workload of the teachers they are meant to help. It was found in both studies that robots could be surprisingly helpful as assistants in lessons and that if they were not meant to lead lessons they could act more on a personal level with students as tools that can be used to engage children in practicing their lessons and skills. This study involves a robot assistant as it introduces itself to the participant.

% personalization and engagement

Without personalization, interaction with children over time can lead to diminishing returns for the effectiveness of a robot teaching assistant~\cite{leite2014empathic}, as was observed in a study that was conducted over 5 weeks. To combat the loss of engagement over time, the robot would interact with children using a model of empathy in which the robot would verbally console the child after a mistake; this engagement model showed significantly higher engagement with the robot over time. Belpaeme et al. presented an extensive review of in-classroom HRI studies, and showed that robots that were considered social actors and whose behavior was modeled accordingly achieved better results in terms of engagement and acceptance over time~\cite{belpaeme2018social}. In this study, we measure the effect that a child-drawn robot face has on the perceived social intelligence \cite{barchard2020measuring} of the robot.

% robots in therapy

Social robots have also been explored in therapeutic contexts, where empathetic behavior and expressive design can contribute to improved outcomes. For example, Saerbeck et al. \cite{saerbeck2010} used a robot tutor to help children study for a simple language learning exam and found that the group which had a tutor that was supportive and empathetic led to better learning outcomes for the students. Short et al. investigated the development of relationships between an expressive robot and students in a classroom setting~\cite{short2014}. A specialized use case for social robots, which might provide some inspiration, is a robot assist therapists for children with autism spectrum disorders (ASD). An extensive review of work in this space notes that primarily robots are used to teach facial expressions through mimicry, and to facilitate collaboration~\cite{kr2022use}. Another review focuses on the application of robots in actual clinical therapy sessions and notes that the research in the area focuses on the design of the robot and the acceptability of the robot, however, many did not try to utilize clinical methods, and were not generalizable to be used outside of their research~\cite{diehl2012clinical}. In this study, we focus on creating more approachable and personalized robots which could also be used as social actors in these clinical spaces.

\begin{figure}[t]
    \centering
    \includegraphics[width=0.50\linewidth]{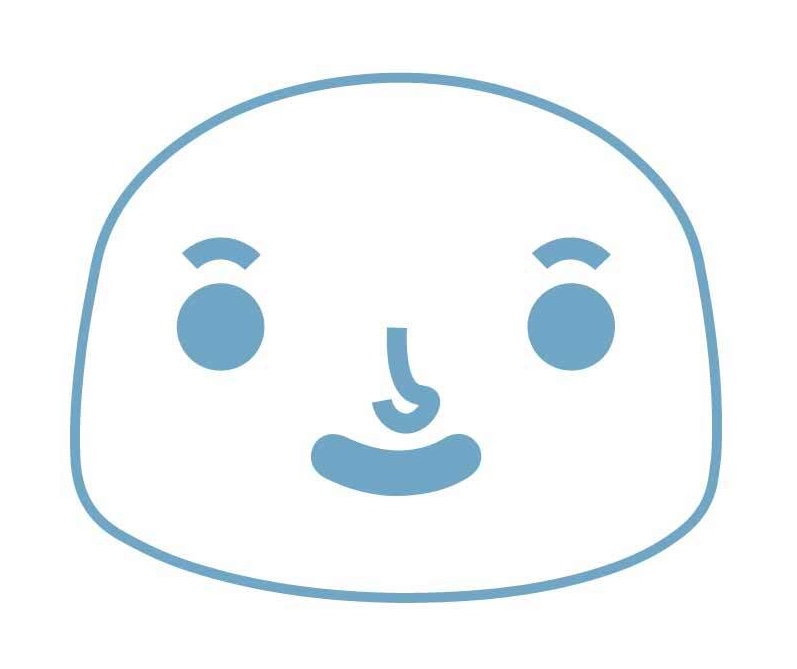}
    \caption{The generic face was designed with features that are traditionally included in existing assistive robots.\cite{kalegina2018}}
    \label{fig:blankrobotgeneric}
\end{figure}

\subsection{Expressiveness and Communication}

Robots employ affective computing to help inform decision-making around expressive behavior and enable robots to better communicate with and work with their human collaborators.  Affective computing is the bridge between these studies and computing with a specific focus on robots~\cite{picard1997affective}.  Current robot studies use Ekman's model of basic emotions (consisting of joy, sadness, disgust, fear, surprise, and anger) as well as Action Units~\cite{ekman1978facial} as a basis for what emotions can be communicated, and how they should be communicated. More recently, these basic emotions have been expanded into 28 emotional naturalistic states~\cite{Cowen2019-xx}. This more nuanced model of emotional states groups expressions with relevant emotions, specifying subgroups to classify expressiveness. These expressive states have been validated, showing that people can recognize these emotions when they are expressed by other people~\cite{buck1972communication}.

For this study, the above is especially helpful as a model because children have been shown to learn to recognize facial expressions early in their development~\cite{camras1985socialization}. While this study does not specifically focus on communicating emotion, these are relevant studies into what expressive affordances a robot face requires, as described in the next section. These serve as a basis for the purpose that facial expression serves in communication; in this study we are evaluating the effect that using a child-drawn face has on the social capabilities of a robot which can serve as a basis for more effective communication in future designs.

\subsection{Robot Faces and Personalization}

Robot faces have been identified as a key problem in designing a social agent, especially for one meant to communicate with children. Chesher et al.~\cite{chesher2021science}classify robot faces into five different categories: anthropomorphic, blank, symbolic, tech, and screen. When it comes to expression and in particular emotional expression, only anthropomorphic, symbolic and screen faces are of interest as the others cannot express emotion with only their face. In this study, we utilized a screen face in the form of a tablet attached to the end of a robotic arm. This enabled us to allow the participants to design a face and apply it to the robot without a need to change any physical components of the robot.

Symbolic faces are typically used in spaces for education, such as the Nao robot, the iCat, and the Keepon~\cite{belpaeme2018social}. This is due to their ability to be expressive through motion as well as through the motion of their neck or body. Bennet et al. even showed that the abstraction of the face could be taken even further by having study participants identify the expressed emotions of a robot face (except for disgust) which consisted of only a mouth, eyes, and eyebrows~\cite{bennett2014deriving}. Conversely, Breazeal et al. designed a much more complicated robot face for expressing emotion, which was able to express all emotions such that they could be accurately identified called Kismet~\cite{breazeal2003emotion}. In this study, we instructed the participants to focus on the key features identified by Bennet et al. rather than a more complicated face, in order to make drawing faces easier for children without losing the expressiveness of the face. Our face was displayed on a screen, yet it retained expressive properties similar to Bennet et al.

Personalization has been shown to be a key factor in increasing the acceptability of robots in society. Personalization has been studied in the vein of interactions with people (i.e., what the robot should be saying or doing based on a person's culture, gender, prior experience with robots, etc.). Ligthart et al. proposed a system of utilizing a through-line narrative as well as remembered portions of prior conversations with users to simulate a personalized continued conversation over the course of multiple sessions with participants~\cite{ligthart2022memorybased}. They showed through their accompanying validation study that this system increased continued engagement over time with the user whereas the control group lost interest in interacting with the robot more quickly over time.

In addition to promoting engagement, personalization can help with the acceptance of a robot. Gasteiger et al. provide a meta-review of research into personalization with respect to cultural differences and showed that personalization is a necessary step forward in ingratiating robots into different societies~\cite{gasteiger2021optimizing}. Personalization is best measured in long-term studies in HRI in order to better understand how the addition of a robot in these spaces can help over a long period of time rather than studying short-term encounters~\cite{irfan_24}. Our study instead focuses on allowing the participants to create their own robot face which provides a elucidates a different perspective on personalization than has been the focus in prior work. To our knowledge there has been no research on enabling the user to design their own robot face and no prior study on how this would effect their perception of that robot.

\subsection{Co-design With Children}

Children as collaborative designers for SARs can serve in the following roles: users, testers, informants, design partners, co-researchers, and protagonists~\cite{10.1145/3434073.3444650}. Neerincx et al.~\cite{NEERINCX2023100615} use a workshop pipeline to involve children, who would be end-users, eliciting a child's creativity to inform robot behaviors and functionality. Alves-Oliveira et al.~\cite{10.1145/3078072.3084304} similarly enable children to help design a robot by letting them create stories, utilizing blocks that they would give personalities and animate. In both cases, the robot was not completed during the course of the study, and therefore the children could only be design partners.

Our approach involves children as design partners through their contributions of their personalized drawings. Because the children also see the product of their work and evaluate the difference between a generic face and their drawings, they also serve as a users and informants. 

\subsection{Perceived Social Intelligence (PSI)}

The PSI measure is used to analyze how a robot's behavior effects its perceived social intelligence, compare different robot embodiment considerations, consider the effect of context on perceived social intelligence, examine how PSI influence task objectives, and analyze the relationship between social intelligence and personality~\cite{barchard2020measuring}. In our study, we use the PSI as a tool to measure the difference between two different physical considerations for a SAR. The subscales allowed us to clearly label the aspects of social interaction that could be relevant in spaces where a CRI assistive robot would be implemented in. The subscales considered for this study were \textit{Identifies Humans} [\textbf{IH}], \textit{Friendly} [\textbf{FRD}], \textit{Identifies Individuals} [\textbf{II}], \textit{Trustworthy} [\textbf{TRU}], \textit{Caring} [\textbf{CAR}], \textit{Perceives Emotion} [\textbf{PE}], \textit{Social Competence} [\textbf{SOC}], \textit{Conceited} [\textbf{CON}], \textit{Rude} [\textbf{RUD}], and \textit{Hostile} [\textbf{HOS}]. This selection enables a structured evaluation of social responses, integrating multiple dimensions of social perception into a coherent measurement approach. To make the questions appropriate for the participants, the language was modified for the age group.

\begin{figure}[t]
    \centering
    \includegraphics[width=\linewidth]{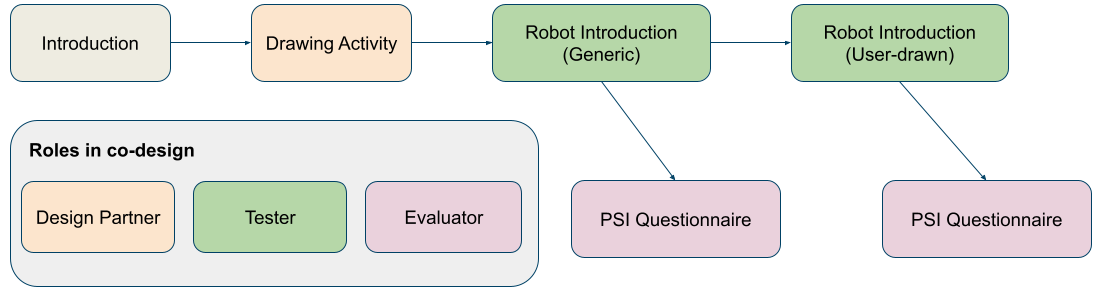}
    \caption{Participants served three different roles for the duration of the study. Each child was a design partner, tester, and evaluator of the system.}
    \label{fig:study_phases}
\end{figure}

\begin{figure*}[h]
    \centering
    \includegraphics[width=\textwidth]{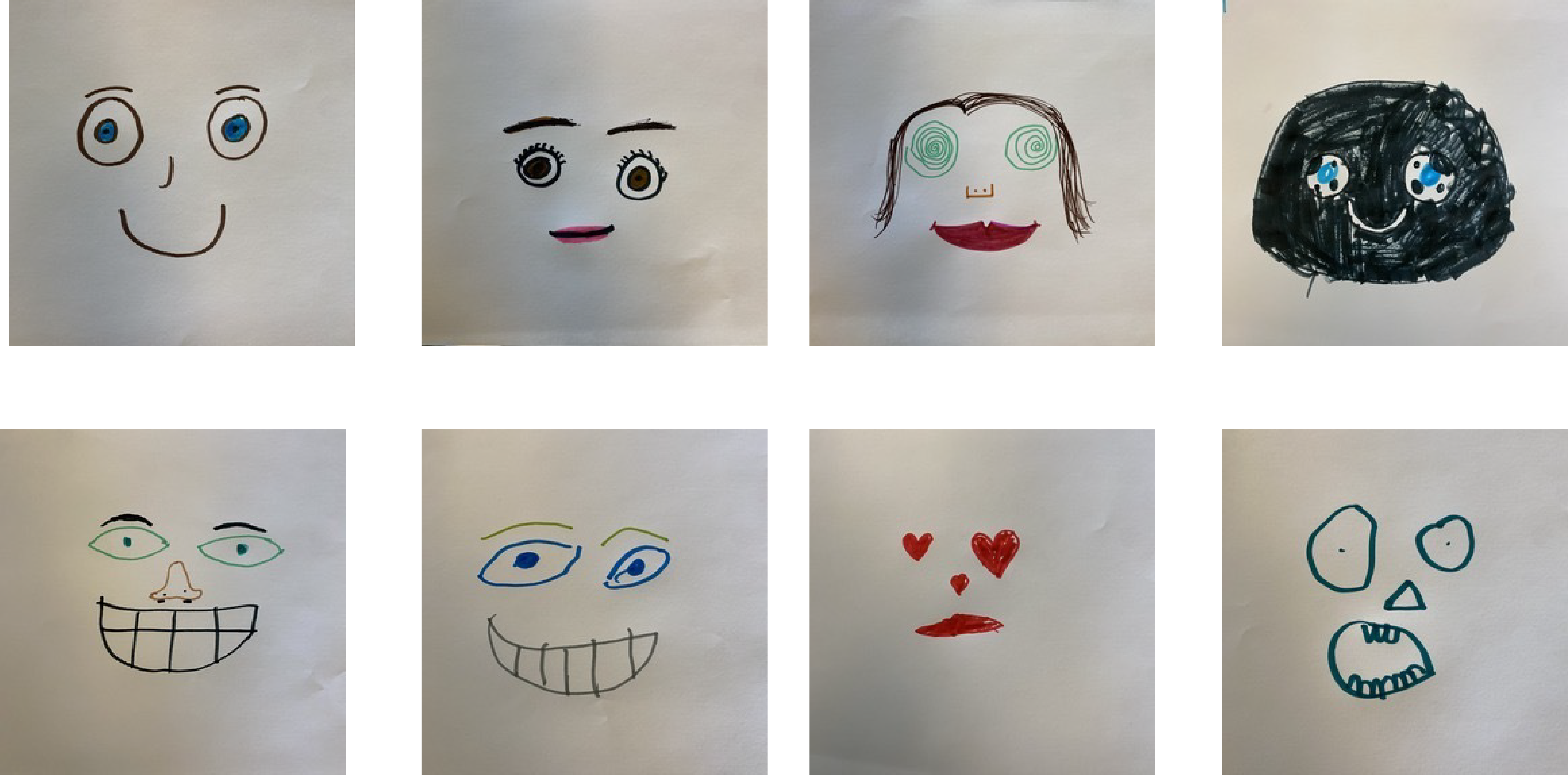}
    \caption{Participants drew faces in varying styles. Many stylistic choices were unique to each participant which indicated to each participant's personality reflected somewhat into the drawings. Some common features in the drawings include: circle eyes, pupils of different sizes, and smiling mouths.}
    \label{fig:faces_grid}
\end{figure*}

\section{Methods} \label{Methods}

The purpose of this study was to find out if using user-designed faces for robot agents would produce positive responses from children. We compared the perceived social intelligence of a generic robot face with a child-drawn face. This was a single-session within-subject study. 

\subsection{Research Questions}

We examine the following research question:

\begin{itemize}
    \item {\bf RQ1}: How does a child-drawn digital-faced robot compare to a generic robot face when rating the social proficiency of the robot?
\end{itemize}

We will measure RQ1 through Likert scale questions derived from the PSI scale. See \ref{Questions} for a detailed description of the questions in the survey.  Our hypothesis for this research questions is:
\begin{itemize}
    \item {\bf H1}: Robot agents with child-drawn faces will be perceived with \textit{higher social intelligence by the user who designed the agent}.
\end{itemize}

\subsection{Study Design} \label{StudyDesign}

\subsubsection{Experiment Setup}
The activity was advertised to all eligible children in the local STEM museum. We conducted the experiment on a rolling basis, where a room was split into a craft/drawing space and a space to interact with the robot. The drawing activity was contextualized with an introduction to different robot embodiments as a primer for the children drawing their own faces. This primer did not include any examples of robot faces to prevent bias for their own drawings. A researcher guided the children through the activity, asking them to verbally brainstorm what they believed would be appropriate to include in a robot face. Examples of these child-drawn faces are seen in Figure~\ref{fig:faces_grid}.

At the completion of the co-design drawing activity, participants were instructed to sit in front of a robot with a digital screen for a short introduction interaction. Following a scripted dialogue from the robot, the participants were introduced to the agent. The interaction was divided into two sections: the child-drawn agent and a generic agent. After each section, the experimenter verbally administered survey questions. This allowed for any clarifications from the participant to maintain response accuracy. The order of these sections was randomized between participants. For the child-drawn agent, the drawings were displayed on the digital face of the robot generated using the method described in Section \ref{Animation}. At the end of both interactions, the participant was asked their overall preference between the two agents with a question asking "did you prefer the generic face or the drawn face?". The flowchart of all phases of the study is seen in Figure~\ref{fig:study_phases}

\subsection{Participant Recruitment}

The study population included children in 1$^{st}$ grade to 6$^{th}$ grade. The median age of the participants was 9 years old. Parent or guardian permission was attained prior to participation and assent from participants were obtained before starting the activity. The study was conducted in-the-wild at a local children's museum. The duration for participation was approximately 20-30 minutes.

We recruited $37$ participants via fliers and in person advertisement in an IRB-approved\footnote{IRBNet ID: 2236279-1} study (an additional 12 potential participants did not meet the age criteria and were not included in the study). In all, $17$ females and $20$ males participated in the study.

\subsubsection{Questionnaire} \label{Questions}
The questionnaire consisted of questions from the perceived social intelligence (PSI) instrument with the question language modified slightly to be age-appropriate. The questionnaire was administered after each face was shown on the robot embodiment. Because the population of the study included younger children, participants were given the option of dictation or completing the survey on their own.

Each question covers a subscale of the PSI. These questions fall under the scales that are listed with the statement. To keep the evaluation simple and explainable to the participants, the statements were represented as an overarching statement to summarize the subscale.  

The questions in the survey were measured on a 5-point Likert scale based on participant agreement with the following statements:

\begin{enumerate}
    \item The robot notices you [\textbf{IH}]
    \item The robot enjoys meeting people [\textbf{FRD}]
    \item The robot recognizes people [\textbf{II}]
    \item The robot is trustworthy [\textbf{TRU}]
    \item The robot cares about others [\textbf{CAR}]
    \item The robot recognizes your emotions [\textbf{PE}]
    \item The robot is a smart friend [\textbf{SOC}]
    \item The robot thinks it is better than everyone else [\textbf{CON}]
    \item The robot is impolite [\textbf{RUD}]
    \item The robot wants to hurt people [\textbf{HST}]
\end{enumerate}

\subsection{Generic Face Design}
The face design for the generic face includes eyebrows, eyes, a nose, and a mouth. This design takes inspiration from a simple digital robot face with minimal features \cite{kalegina2018}. The color of the features was also chosen to be a neutral, powder blue seen in Figure~\ref{fig:blankrobotgeneric}.

\subsection{Child-Drawn Face Animation} \label{Animation}

\begin{figure}[!t]
  % \vspace{-2em}
    \centering
    \captionsetup[subfigure]{justification=centering}
        \subfloat[Drawn Face]{
          \includegraphics[width=0.21\textwidth]{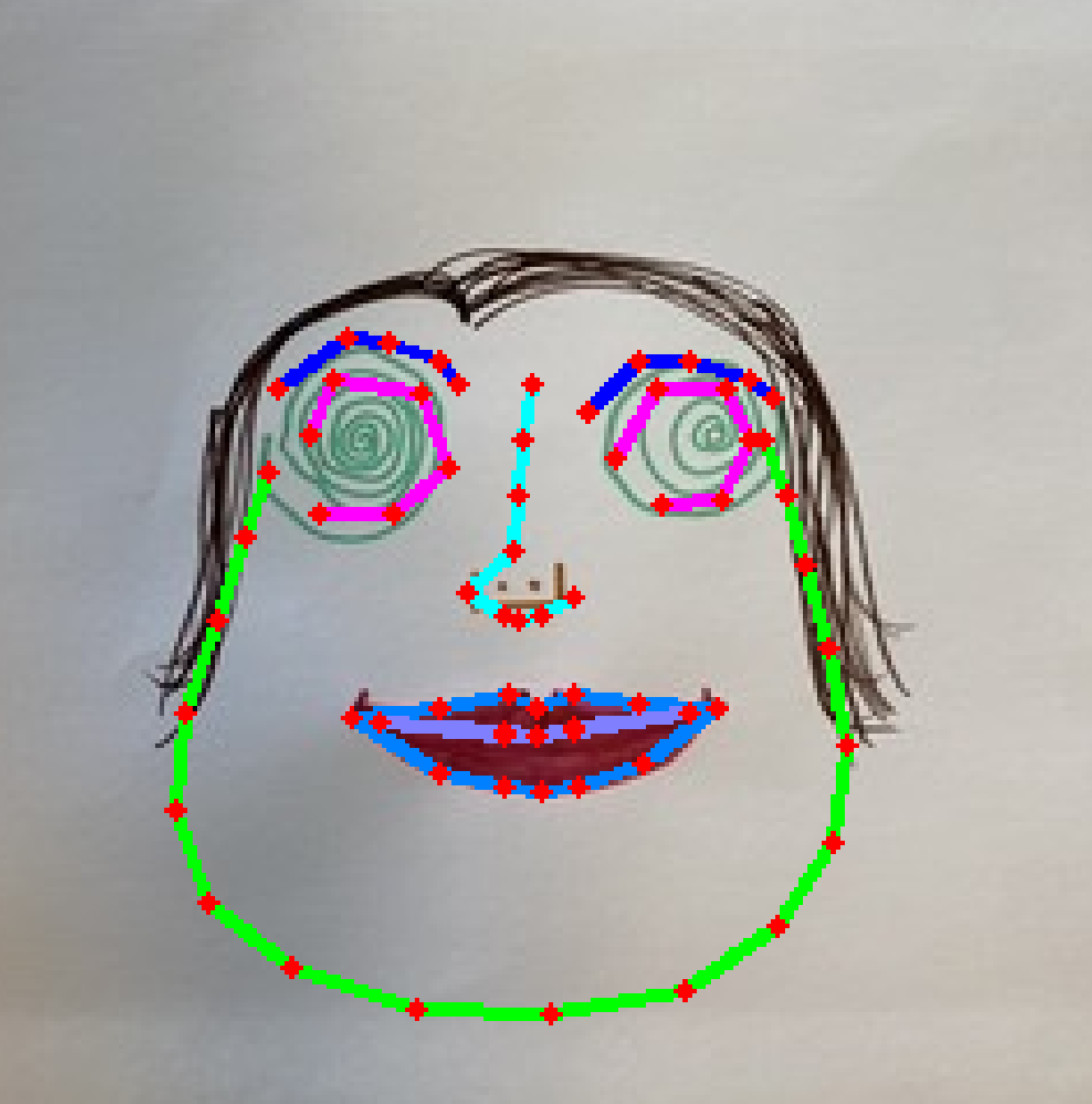}
          \label{landmarks_drawn}
          }
        \subfloat[Key Features for Animation]{
          \includegraphics[width=0.236\textwidth]{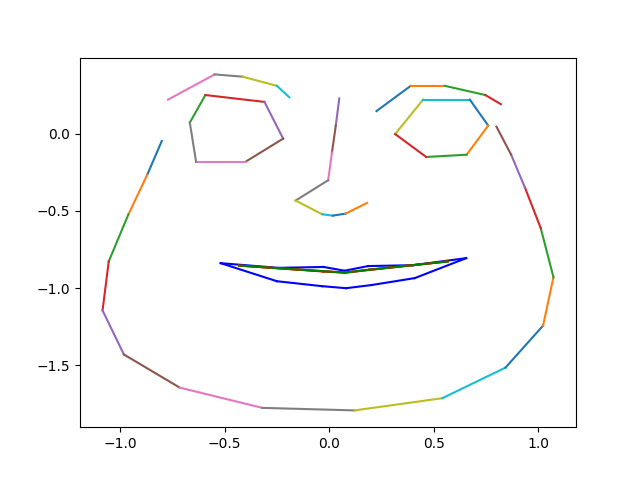}
          \label{landmarks_generated}
          }    
    \caption{Drawn faces were fixed with facial landmarks relating to features including eyebrows, eyes, nose, mouth, and chin. A generated animation space would then take the placed landmarks and use it to animate the drawings.}
    \label{fig:visualization}
\end{figure}

Animating static drawings drawn by each child allowed us to observe the effects of personalization. We converted participants' drawings of a robot face to a video visible on the robot's screen as quickly as possible. We utilized a system inspired by related face animation work, which used custom weights for dlib's~\cite{Zhou_2020} standard facial recognition landmarks as seen in Figure~\ref{fig:visualization} in order to estimate facial features on the participant's drawings. After the landmarks were identified and hand-corrected, we generated an animation based on those drawings and synced to a pre-recorded audio clip.

\section{Results} \label{Results}

Each PSI sub-scale was evaluated for significant differences in the responses between the two groups. The child-drawn faces were also observed for what features were included and how they were drawn.

\begin{figure*}[h]
    \centering
    \includegraphics[width=\textwidth]{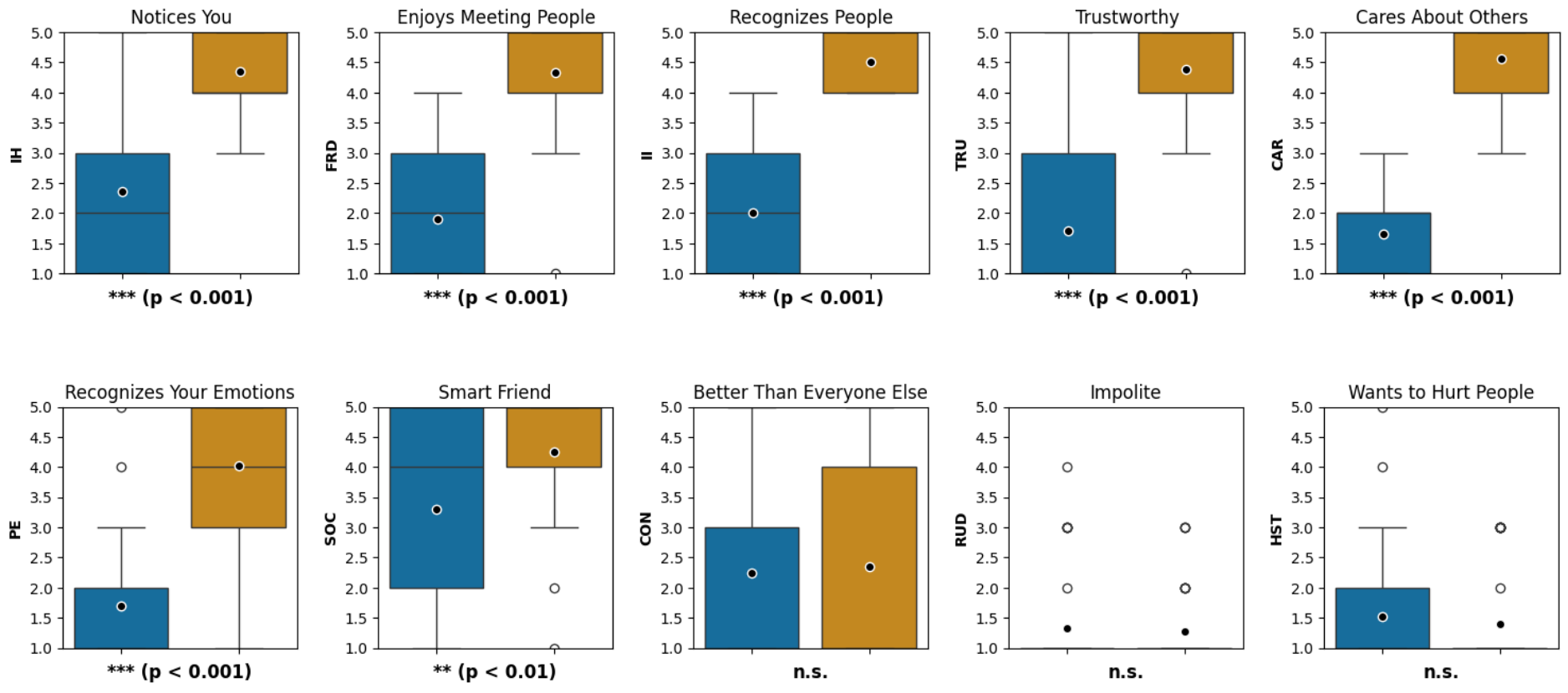}
    \caption{The scores of the child-drawn face is higher than the scores of the generic face seen in the yellow colored plots. The generic face has scores that span a larger range with many questions having ranges that include the entire set shown in the dark blue colored plots. Question wording can be found in Section~\ref{Questions}.}
    \label{fig:likert_split}
\end{figure*}

\subsection{Perceived Social Intelligence}

In this section, we present the results for RQ1 relating to how child-drawn faces compare to generic faces.

A Shapiro-Wilk test was used to verify the normality of the response distribution of the survey questions. From this analysis, the p-value for each of the questions was (\textit{\textbf{p \textless 0.05}}) showing that an ANOVA test would be insufficient as the data were not normally distributed.  A Kruskal-Wallis test was used to compare responses between the child-drawn and generic face conditions, showing a significantly higher perception of social intelligence for the child-drawn face than the generic face (\textit{\textbf{p \textless 0.001}}).

\begin{figure}[h]
    \centering
    \includegraphics[width=0.75\linewidth]{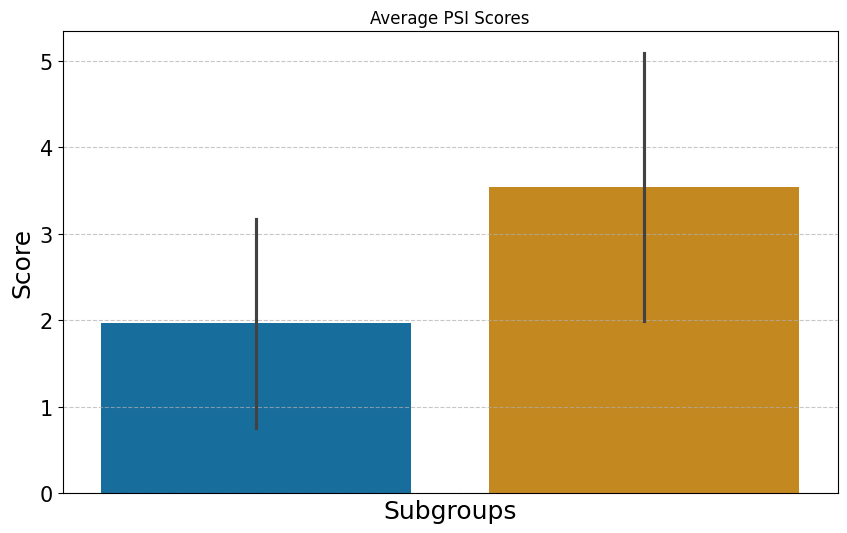}
    \caption{The PSI average along all of the questions for the two faces show the higher scores for the child-drawn face over the generic face. The lighter orange color represents the child drawn faces and the darker blue represents the generic faces.}
    \label{fig:likert_overall}
\end{figure}

To examine the survey answers in more detail, a post hoc analysis of each question was also performed. The Mann-Whitney U statistic was used to find which scales had significant differences in rankings. The first survey question, \textit{``The robot notices you"} showed that participants felt that the robot noticed them more (IH) with a child-drawn face than a generic face with (\textit{p \textless 0.001}). The next question, \textit{``The robot enjoys meeting people''} showed that participants felt that the robot enjoyed meeting people more (FRD) with the child-drawn faces over the generic face and yielded a p-value of (\textit{p \textless 0.001}). The third question, \textit{``The robot recognizes people"} showed that participants felt that the robot recognized people (II) more than with a generic face, and yielded a p-value of (\textit{p \textless 0.001}). The fourth question, \textit{``The robot is trustworthy"} showed that participants felt that the child-drawn robot face was more trustworthy (TRU) than the generic face and yielded a p-value of (\textit{p \textless 0.001}). The fifth question, \textit{``The robot cares about others"} showed that participants felt that the child-drawn robot face was more caring about others (CAR) when compared to the generic face and yielded a p-value of (\textit{p \textless 0.001}). The sixth question \textit{``The robot recognizes your emotions"} showed that participants felt that the robot recognized their emotions more (PE) with a child-drawn face than a generic face, with (\textit{p \textless 0.001}). The seventh question, \textit{``The robot is a smart friend"} showed that participants felt that the robot was a smart fried (SOC) with the child-drawn robot face and the generic face with (\textit{p~\textless~0.01}). The eighth question, \textit{``The robot thinks it is better than everyone else"} showed that participants felt that the robot was better than everyone else (CON) with a child-drawn face and a generic face with (\textit{p \textgreater 0.05}). The ninth question, \textit{``The robot is impolite"} showed no significance in the way participants felt about the robot being impolite (RUD) with a child-drawn face or a generic face with (\textit{p \textgreater 0.05}). Finally, the tenth question, \textit{``The robot wants to hurt people"} showed no significance in the way that children felt about the robot hurting people (HST) with a child-drawn face and a generic face, with (\textit{p \textgreater 0.05}). The average performance of all the questions are summarized in Figure~\ref{fig:likert_overall}. A summary of the significant relations are visualized in Figure~\ref{fig:likert_split}.

Conducting a power analysis for each scale yielded a power~\textgreater~0.8 for the following scales: IH, FRD, II, TRU, CAR, PE. The subscale SOC had a power~\textgreater~0.8. The last three subscales CON, RUD, and HST had a power \textless 0.8 that further contributed to the previous results.

\subsection{Commonly-Drawn Facial Features}

In addition to the PSI, we observed the occurrence of different face features and how they were drawn.

Eyes were included in all the faces drawn, but varied in what shape, how big, and how they were drawn. Some participants included circle shaped eyes, but not all circle eyes were drawn with pupils. Twenty-Six of the 37 participants included pupils in their eyes. Twelve of the 37 participants drew eyes that were not circles (ovals, lines, etc.). 18 out of the 37 participants drew eyebrows to accompany their eyes. 

There were two distinct groups for the way that mouths were drawn. Children drew line mouths or open mouths in the shape of a smile. Some features included were teeth and tongues. Nineteen of the 37 participants drew open/smile-like mouths. Twelve participants drew line mouths in the shape of a smile. Noses were varied among all participants and some drawings omitted noses entirely. Noses were drawn in a circle, in a small line, or another shape, but no style was heavily adopted. 

The observed features provided insight into the preferences of the children when drawing facial features. Participants demonstrated preference in drawing the main facial features rather than including features such as the nose and eyebrows. This provides a sense of the facial features that children can most easily recall in a personalized robot face. Examples of these child-drawn faces are seen in Figure~\ref{fig:faces_grid}.

\section{Discussion} \label{Discussion}

We interpret these findings through three lenses: cognitive factors influencing preference, implications for HRI design, and potential impacts in educational and therapeutic settings.

% paragraph on preference
The majority of participants showed preference toward the child-drawn robot faces, and some expressed a sense of attachment even after the brief time that they had with the robot. This phenomenon may be rooted in cognitive factors like the endowment effect~\cite{Reb_Connolly_2007} that states that humans have the tendency to value things that we have a part in creating. The familiarity and personalized nature of the child-drawn faces likely elicit a positive emotional response as opposed to the generic face. Participants may have projected a sense of ownership making it feel more trustworthy, friendly, and emotionally expressive. This aligns with findings in personalization literature, where users develop emotional bonds with customized devices and interfaces~\cite{Blom01092003}. On the other hand, the generic robot might lack this sense of ownership and familiarity. This could inform the lower scores that the generic face received for trust and sociability.

% implications for design
The findings presented in this paper have important implications for the designs of SARs in various spaces. The significant differences across seven of the ten questions evaluated indicate that child-drawn faces can significantly enhance how a SAR is perceived. Aesthetic personalization can strengthen a SAR's social presence and even perceived competence. This suggests that implementing a user-in-the-loop approach with children co-creating robot features can develop deeper engagement and relational connection. While generic designs can be scalable, they may lack affordances needed for impactful interactions in social and assistive roles.

% impacts on education settings

In educational spaces where a robot serves the role of a tutor, a practice tool, or a companion, personalized faces could also increase motivation and attention through the preferences of a single child or a group of students. Creating a sense of personal connection between a student and an assistive robot could decrease the emotional distance for a robot agent in this space and increase positive learning experiences and outcomes. In comparison, generic robot faces may fail to elicit the level of engagement needed in contexts that rely on emotional resonance. Integrating a child-drawn robot face could be a meaningful step toward improving outcomes in both educational and therapeutic interventions.

\section{Limitations} \label{Limitations}

While the results of this design activity show a significant difference in child-drawn and generic robot face agents, there are limitations that should be addressed in future iterations of this research. 

Studying long-term effects of interaction with personalized agents should be a focus for future research. The work done in this study aimed to find if personalization in the sense of drawing your own robot would perform better in comparison to a non-personalized robot agent. However, the ultimate goal of this approach is to implement a similar system for long-term interventions. Exploring long-term interactions can reveal how relationships with customized systems evolve over time.

Finally, these results show a distinct user {\it preference} for the child-drawn face, there is no evaluation yet if such personalization results in a robot that promotes better clinical or educational {\it outcomes}, such questions were out of scope for this study. However, since perception of social intelligence can relate to trust, competence, and politeness, there is some evidence that such factors might lead to improved outcomes.

\section{Conclusion} \label{Conclusion}

This paper shows a design activity where children could personalize an animated robot face. We examined which features were commonly drawn by children. We also examined how children rated child-drawn vs. generic robot faces. The results showed that social perception of a child-drawn robot face was higher than a generic robot face. This showcases the potential of using facial personalization for the design of socially assistive robots. Personalized robot faces can not only increase the trust and likability of an agent but also aid in the adoption of a robot in a therapeutic or educational setting. 

\section*{Acknowledgment}

This material is based upon work supported under the AI Research Institutes program by National Science Foundation and the Institute of Education Sciences, U.S. Department of Education through Award \# 2229873 - AI Institute for Transforming Education for Children with Speech and Language Processing Challenges. Any opinions, findings and conclusions or recommendations expressed in this material are those of the author(s) and do not necessarily reflect the views of the National Science Foundation, the Institute of Education Sciences, or the U.S. Department of Education. This work was supported by the National Science Foundation (IIS 2150394). 

This work was made possible by the support of the The Terry Lee Wells Nevada Discovery Museum in Reno, NV, USA. We would also like to thank Sachin Parajuli for his help during data collection.

\bibliographystyle{IEEEtran}
\bibliography{conference_101719}

\end{document}